\documentclass[a4paper]{article}

\usepackage{INTERSPEECH2020}

\newsavebox{\imagebox}
\usepackage[shortlabels]{enumitem}
\usepackage{url}
\usepackage{cite}

\usepackage[flushmargin,ragged]{footmisc} 
\makeatletter
\def\blfootnote{\gdef\@thefnmark{}\@footnotetext}
\makeatother

\title{Dynamic Prosody Generation for Speech Synthesis Using Linguistics-Driven Acoustic Embedding Selection}
\name{Shubhi Tyagi, Marco Nicolis, Jonas Rohnke, Thomas Drugman, Jaime Lorenzo-Trueba}
\address{
   Amazon.com, Cambridge, United Kingdom}
\email{\{tshubhi,nicolism,rohnj,drugman,truebaj\}@amazon.com}

\begin{document}

\maketitle
\begin{abstract}
 Recent advances in Text-to-Speech (TTS) have improved quality and naturalness to near-human capabilities. But something which is still lacking in order to achieve human-like communication is the dynamic variations and adaptability of human speech in more complex scenarios. This work attempts to solve the problem of achieving a more dynamic and natural intonation in TTS systems, particularly for stylistic speech such as the newscaster speaking style. We propose a novel way of exploiting linguistic information in VAE systems to drive dynamic prosody generation. We analyze the contribution of both semantic and syntactic features. Our results show that the approach improves the prosody and naturalness for complex utterances as well as in Long Form Reading (LFR).
\end{abstract}

\section{Introduction}
\vspace{-1mm}
\blfootnote{Samples will be available at: \url{https://www.amazon.science/blog/more-natural-prosody-for-synthesized-speech}}
\label{sec:intro}
Recent advances in TTS have significantly improved the naturalness of synthetic speech ~\cite{shen2018natural,prenger2019waveglow,lorenzo2019towards,hsu2018hierarchical}. 
One aspect that most systems are still lacking is the natural variability of human speech, which is being observed as one of the reasons why the cognitive load of synthetic speech is higher than that of humans \cite{govender2018using}. This is something that variational models such as those based on Variational Auto-Encoding (VAE) \cite{hsu2018hierarchical,akuzawa2018expressive} attempt to solve by exploiting the sampling capabilities of the acoustic embedding space at inference time. 

Despite the advantages that VAE-based inference brings, it also suffers from the limitation that to synthesize a sample, one has to select an appropriate acoustic embedding for it, which can be challenging. A possible solution to this is to remove the selection process and consistently use a centroid to represent speech. This provides reliable acoustic representations but it suffers from the monotonicity problem of conventional TTS. On the other hand random sampling from acoustic space would reduce the monotonicity but can lead to erraticness for longer texts.. Finally, one can consider text-based selection or prediction, as done in this research.

The tight relationship between syntactic constituent structure and prosody is well known \cite{SyntaxProsody,ProsodyReview}. In the traditional Natural Language Processing (NLP) pipeline, constituency parsing produces full syntactic trees. Recent relevant work exploring the advantages of exploiting syntactic information for TTS can be seen in \cite{Guo2019,Aubin2019}. Those studies, without any explicit acoustic pairing to the linguistic information, inject a number of curated features concatenated to the phonetic sequence as a way of informing the TTS system.

More recent approaches based on Contextual Word Embedding (CWE) suggest that CWE are largely able to implicitly represent the classic NLP pipeline\cite{tenney}, while still retaining the ability to model lexical semantics \cite{mikolov2013distributed}. However simply plugging such embeddings as a feature during synthesis has shown not to perform well \cite{CWEEffects}. 

On the other hand, the present study explores more appropriate ways to exploit linguistic information specifically for VAE based synthesis. We do so by driving the acoustic embedding selection to guide prosodic contour rather than using them as additional model features. 

An exploration of how to use linguistics as a way of predicting adequate acoustic embeddings can be seen in \cite{stanton2018predicting}, where the authors explore the path of predicting an adequate embedding by informing the system with a set of linguistic and semantic information. This work predicts a point in a high-dimensional space by making use of sparse input information (which is a challenging task and potentially vulnerable to training-domain dependencies). This work differs as we use the linguistic information to predict the most similar embedding from our training set, reducing the complexity of the task significantly.

The main contributions of this work are:
\emph{i)} we present a novel approach for linguistically informed acoustic embedding selection during VAE synthesis;
\emph{ii)} we compare the proposed approach with simply including linguistic information as additional features in VAE based TTS; \emph{iii)} we demonstrate that this embedding selection approach improves the overall TTS quality along with prosody in complex scenarios and LFR; \emph{iv)} Finally, we compare the improvements achieved by exploiting syntactic information in contrast with those brought by CWE.

\section{Proposed Approach}
\vspace{-1mm}
\label{sec:propsosedsystems}
We explore the following two hypotheses in our experiments:
\emph{(i)} linguistic information has been used sub optimally in TTS synthesis. Using this information to drive the embedding selection in the VAE system will result in improved prosodic quality as compared to using it as additional features; \emph{(ii)} 
in some scenarios, syntax will be able to generalize better than CWE.

The objective of this work is to exploit sentence-level prosody variations available in the training dataset while synthesizing speech for the test sentence. The steps executed in this proposed approach are:
\emph{(i)} Generate suitable vector representations containing linguistic information for all the sentences in the train and test sets,
\emph{(ii)} Measure the similarity of the test sentence with each of the sentences in the train set. We do so by using cosine similarity between the vector representations as done in \cite{semeval} to evaluate \textit{linguistic similarity} (LS),
\emph{(iii)} Choose the acoustic embedding of the train sentence which gives the highest similarity with the test sentence,
\emph{(iv)} Synthesize speech from VAE-based inference using this acoustic embedding

\subsection{Systems}
\vspace{-1mm}
\label{sec:systems}
We experiment with three different systems for generating vector representations of the sentences, which allow us to explore the impact of both syntax and semantics on the overall quality of speech synthesis. These representations are used to select a sentence level acoustic embedding from the training set.

\subsubsection{Syntactic}
 \vspace{-2mm}
\label{ssec:synt}
Syntactic representations for sentences like constituency parse trees need to be transformed into vectors in order to be usable in neural TTS models. Some dimensions describing the tree can be transformed into word-based categorical feature like identity of parent and position of word in a phrase \cite{Dall2016}.

The \textit{syntactic distance} between adjacent words is known to be a prosodically relevant numerical source of information which is extracted from constituency trees \cite{shen2017neural}. It is explained by the fact that if many nodes must be traversed to find the first common ancestor, the syntactic distance between adjacent words is high. Large syntactic distances correlate with acoustically relevant events such as phrasing breaks or prosodic resets.

\begin{figure}
  \centering
    \includegraphics[width=0.95\linewidth,height=120pt]{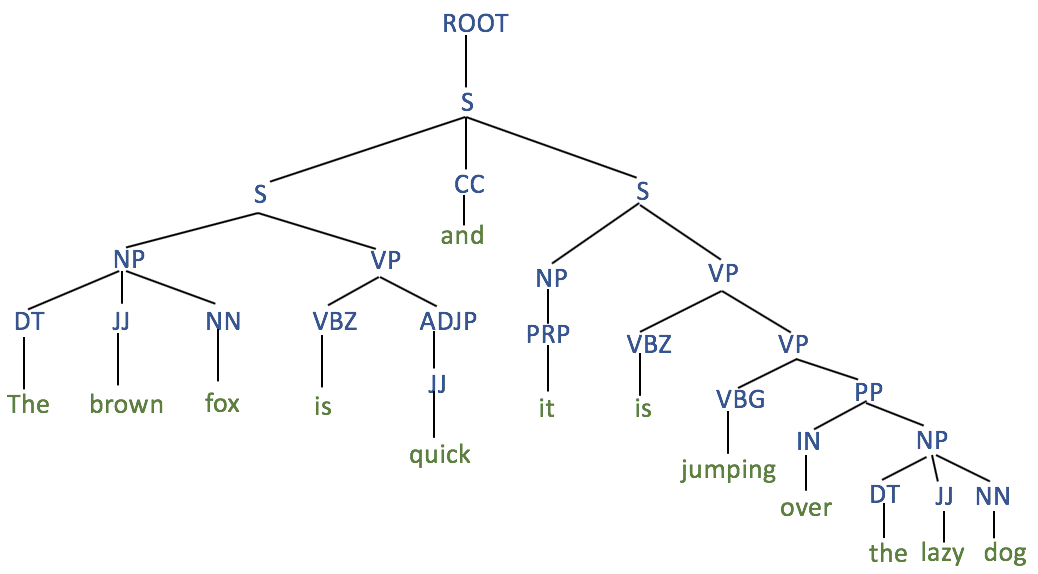}
    	\captionsetup{justification=centering}
    \caption{Constituency parse tree}
    \vspace{-4mm}
\end{figure}

To compute syntactic distance vector representations for sentences, we use the algorithm mentioned in \cite{SyntacticDistance}. That is, for a sentence of \textit{n} tokens, there are \textit{n} corresponding distances which are concatenated together to give a vector of length \textit{n}. The distance for each token is calculated with respect to the previous token. For the first token the distance is always 0.

We see an example in Figure 1:  for the sentence ``The brown fox is quick and it is jumping over the lazy dog", distance vector is \emph{d} = [0 2 1 3 1 8 7 6 5 4 3 2 1]. The completion of the subject noun phrase (after `fox') triggers a prosodic reset, reflected in the distance of 3 between `fox' and `is'. There should also be a more emphasized reset at the end of the first clause, represented by the distance of 8 between `quick' and `and'. 

\subsubsection{BERT}
\vspace{-2mm}
\label{sssec:subsubhead}
To generate CWE we use BERT \cite{BERT}, as it is one of the best performing pre-trained models with state of the art results on a large number of NLP tasks. BERT has also shown to generate strong representations for both syntax and semantics. We use the word representations from the uncased base (12 layer) model without fine-tuning. The sentence level representations are achieved by averaging the second to last hidden layer for each token in the sentence. We do not use `[CLS]' as it acts as an ``aggregate representation" for classification tasks and is not the best choice for quality sentence embeddings vectors.

\subsubsection{BERT Syntactic}
\vspace{-1mm}
\label{sssec:subsubhead}
Even though BERT embeddings capture some aspects of syntactic information along with semantics, we decided to experiment with a system combining the information captured by both of the above mentioned systems. The information from syntactic distances and BERT embeddings cannot be combined at token level to give a single vector representation since both these systems use different tokenization algorithms. Tokenization in BERT is based on the wordpiece algorithm \cite{wordpiece}. On the other hand, tokenization used to generate parse trees is based on morphological considerations rooted in linguistic theory. At inference time, we average the similarity scores obtained by comparing the BERT embeddings and the syntactic distance vectors.

\section{Applications to LFR}
\vspace{-1mm}
\label{sec:lfr}
The systems described in Section~\ref{sec:systems} produce utterances with more varied prosody as compared to the long-term monotonicity of those obtained via centroid-based VAE inference. However, when considering multi-sentence texts, issues of erratic transitions can be introduced. We tackle this  by minimizing the acoustic variation a sentence can have with respect to the previous one, while still minimizing the linguistic distance. We consider the Euclidean distance between the 2D Principal Component Analysis (PCA) projected acoustic embeddings as a measure of acoustic variation, as we observe that the projected space provides us with an acoustically relevant space in which distances can be easily obtained. The 64-dimensional VAE space did not perform as intended, likely because euclidean distances are not being in the non-linear manifold representation. As a result, a sentence may be linguistically the closest match in terms of syntactic distance or CWE, but it will still not be selected if its acoustic embedding is far apart from that of the previous one. 

We modify the similarity evaluation metric used for choosing the closest match from the train set by adding a weighted cost to account for acoustic variation. This approach focuses only on the sentence transitions within a paragraph rather than optimizing the entire acoustic embedding path. This is done as follows:
\emph{(i)} Define the weights for linguistic similarity and acoustic similarity. In this work, the two weights sum up to 1;
\emph{(ii)} The objective is to minimize the following loss for each sentence in the paragraph considering the acoustic embedding chosen for the previous sentence in the paragraph:\\\vspace{-3mm}
\begin{equation}
 Loss = LSW * (1-LS) + (1-LSW) * D
  \label{eq4}
\vspace{-2mm}
\end{equation}
where \textit{LSW = Linguistic Similarity Weight; LS = Linguistic Cosine Similarity between test and train sentence; D = Euclidean distance between the acoustic embedding of the train sentence and the acoustic embedding chosen for the previous sentence}.

We fix D=0 for the first sentence of every paragraph. Thus, this approach is more suitable for cases when the first sentence is generally the carrier sentence, i.e. one which uses a structural template. This is particularly the case for news stories such as the ones considered in this research.

\begin{figure}
	\centering
\includegraphics[width=.80\linewidth,height=90pt] {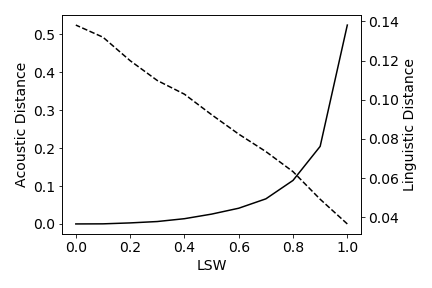}  
	\captionsetup{justification=centering} 
    \caption{Acoustic Distance (solid line) vs Linguistic Distance (dashed line) as a function of LSW across paragraphs}\label{fig:aws}		
    \vspace{-2mm}    
\end{figure}

As LSW decreases, the transitions become smoother. This is not `free': there is a trade-off, as increasing the transition smoothness decreases the linguistic similarity which also reduces the prosodic divergence. Figure~\ref{fig:aws} shows the trade-off when using syntactic distance to evaluate \emph{LS}. Low linguistic distance \textit{(i.e. 1 - LS)} and low acoustic distance are required.

The plot shows that there is a sharp decrease in acoustic distance between LSW of 1.0 and 0.9 but the reduction becomes slower from therein, while the changes in linguistic distance progress in a linear fashion. We informally evaluated the performance of the systems by reducing LSW from 1.0 till 0.7 with a step size of 0.05 in order to look for an optimal balance. At LSW=0.9, the first elbow on acoustic distance curve, there was a significant decrease in the perceived erraticness. 

We performed an internal preference test between the samples generated from LSW=1.0 and LSW=0.9. The results depicted statistical insignificance between the two. However, on individual listening of samples we observed that the paragraphs which comprised of a sentence which shared high linguistic similarity with a more acoustically divergent training sample the preference was given to LSW=0.9. Hence, we chose LSW=0.9 for our LFR evaluations. 

\section{Experimental Protocol}
\vspace{-1mm}
\label{sec:typestyle}
The research questions we attempt to answer are:
\vspace{-1mm}
\begin{enumerate}[start=1,label={\bfseries Q\arabic*:}]
    \item How does using linguistic embeddings as model features impact TTS quality from VAE based systems?
    \vspace{-2mm}
	\item Can linguistics-driven selection of acoustic waveform from the existing dataset lead to improved prosody and naturalness when synthesizing speech ?
	\vspace{-2mm}
	\item How does syntactic selection compare with CWE selection?
	\vspace{-2mm}
	\item Does this approach improve LFR experience as well? 
\end{enumerate}
	\vspace{-1mm}

\subsection{Text-to-Speech System}
\begin{figure}
	\centering
	\includegraphics[width=0.95\linewidth]{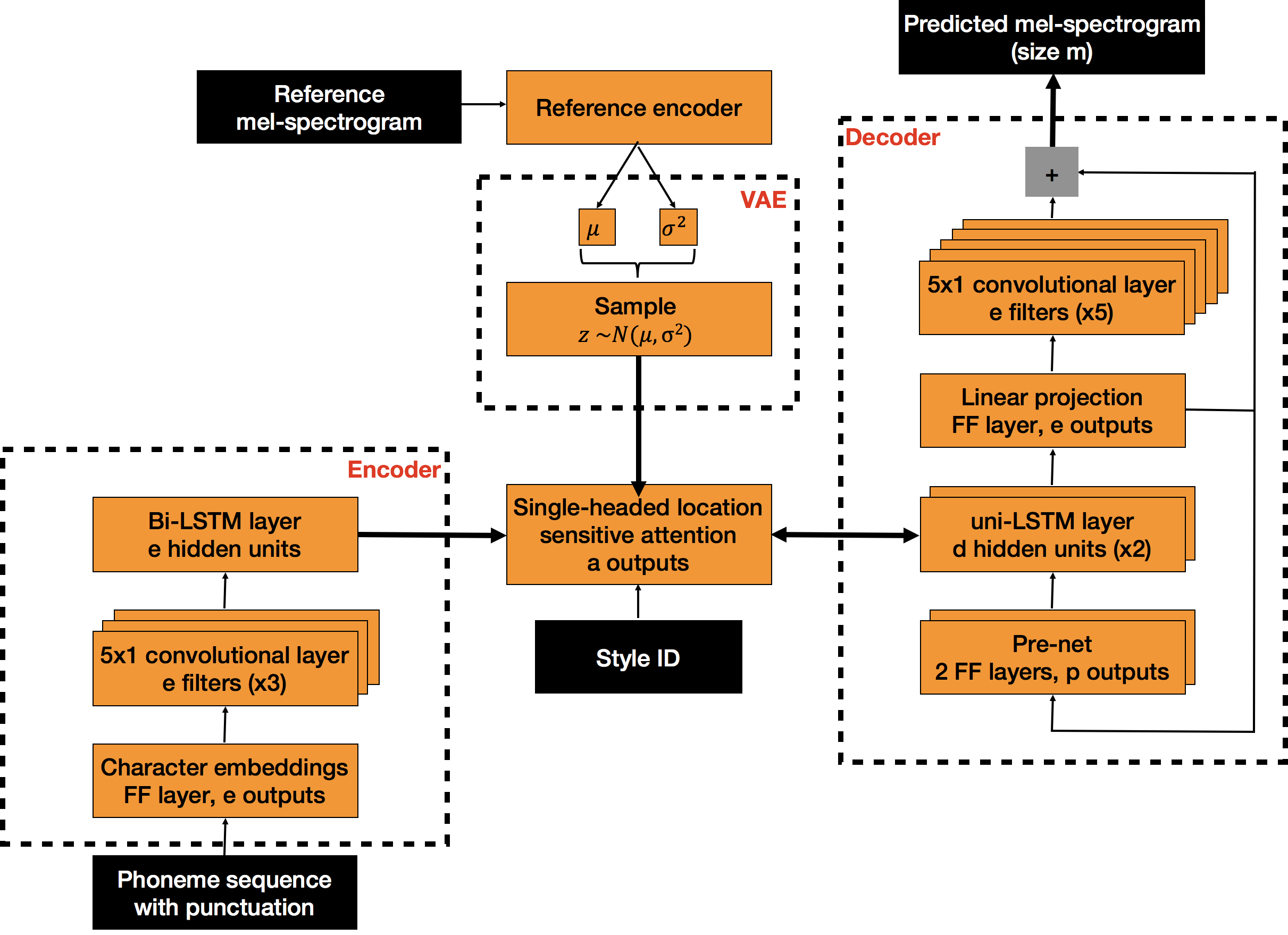}
	\captionsetup{justification=centering}
	\caption{Schematic of the implemented TTS system}
	\label{fig:vae-prosotron-model}
	\vspace{-2mm}
\end{figure}

 \vspace{-2mm}
The evaluated TTS system is a Tacotron-like system \cite{Tacotron} already verified for the newscaster domain. A schematic description can be seen in Figure~\ref{fig:vae-prosotron-model} and a detailed explanation of the baseline system and the training data can be read in~\cite{prateek2019other,latorre2019effect}. Conversion of the produced spectrograms to waveforms is done using the Universal WaveRNN-like model presented in~\cite{lorenzo2019towards}.

For this study, we consider an improved system that replaced the one-hot vector style modeling approach by a VAE-based reference encoder similar to~\cite{akuzawa2018expressive,hsu2018hierarchical}, in which the VAE embedding represents an acoustic encoding of a speech signal, allowing us to drive the prosodic representation of the synthesized text as observed in~\cite{hodari2019using}. Embedding selection at inference time is defined by the approaches introduced in Sections~\ref{sec:systems} and~\ref{sec:lfr}. Embedding dimension is set to 64 to allow for the best convergence without collapsing the KLD loss during training.

\subsection{Datasets}
\vspace{-1mm}
\subsubsection{Training Dataset}
\vspace{-2mm}
\emph{(i)} \textbf{TTS System dataset}:
We trained the TTS system with a mixture of neutral and newscaster style speech for a single speaker in US English. Total of \texttildelow24 hours of training data, split in 20 hours of neutral (22000 utterances) and \texttildelow4 hours of newscaster styled speech (3000 utterances).\\
\emph{(ii)} \textbf{Embedding selection dataset}: 
As the evaluation was carried out on the newscaster style, we restrict our linguistic search space to the utterances associated to the style: 3000 sentences.

\vspace{-2mm}
\subsubsection{Evaluation Dataset}
 \vspace{-2mm}
 \emph{(i)} \textbf{Common Prosody Errors (CPE)}: 
 The dataset on which the baseline Prostron model fails to generate appropriate prosody. This dataset consists of utterances like compound nouns (22\%), “or” questions (9\%), “wh” questions (18\%). This set is further enhanced by sourcing complex utterances (51\%) from \cite{CPEDataset}.\\
 \emph{(ii)} \textbf{LFR}: As demonstrated in \cite{GoogleLFR}, evaluating sentences in isolation does not suffice if we want to evaluate the quality of long-form speech. Thus, for evaluations on LFR we curated a dataset of news samples. The news sentences were concatenated into full news stories, to capture the overall expressive experience. 
 
\subsection{Subjective evaluation}
 \vspace{-2mm}
Our tests are based on MUltiple Stimuli with Hidden Reference and Anchor (MUSHRA)~\cite{recommendation2001method}, but without forcing a system to be rated as 100, and not always considering a reference.

For the CPE dataset, we carried out two tests. One with 10 linguistic experts as listeners, who were asked to rate the appropriateness of the prosody, ignoring speaking style, on a scale from 0 (very inappropriate) to 100 (very appropriate). We chose linguists for this test as prosodic evaluations are complex and require domain specific knowledge. The second test was carried out on 10 crowd-sourced listeners who evaluated the naturalness of the speech from 0 to 100. In both tests each listener was asked to rate 28 different screens, with 4 randomly ordered samples per screen for a total of 112 samples. The 4  systems were the 3 proposed ones and the centroid-based VAE inference was fixed as the baseline.

For LFR it's difficult to get consistent scoring while evaluating prosody, as one needs to remember the entire context. Also, there's no canonically ``correct" rendition of a paragraph. Thus, we conducted a crowd-sourced evaluation only for naturalness where the listeners were asked to assess the suitability of newscaster style on a scale from 0 (completely unsuitable) to 100 (completely adequate). Each listener was presented with 51 news stories, each playing one of the 5 systems including the original recordings as a top anchor, the centroid-based VAE as baseline and the 3 proposed linguistics-driven acoustic embedding selection systems.

All of our listeners, regardless of linguistic knowledge were native US English speakers.

\section{Results}
\vspace{-1mm}
\label{sec:print}
For \emph{\textbf{Q1}}, we ran another internal preference test on the CPE dataset between the centroid baseline model with and without BERT CWEs as additional features. We chose CWE for this test as they are believed to capture both semantics and syntactics. The BERT based model performed significantly worse (p\textless\textless0.01) than the centroid baseline as we believe  it was unable to generalise to unseen scenarios due to the sparsity of the linguistic-acoustic mapping space at utterance level. Hence, we eliminate this system from further evaluations.

\begin{figure}
	\centering	\includegraphics[width=0.95\linewidth]{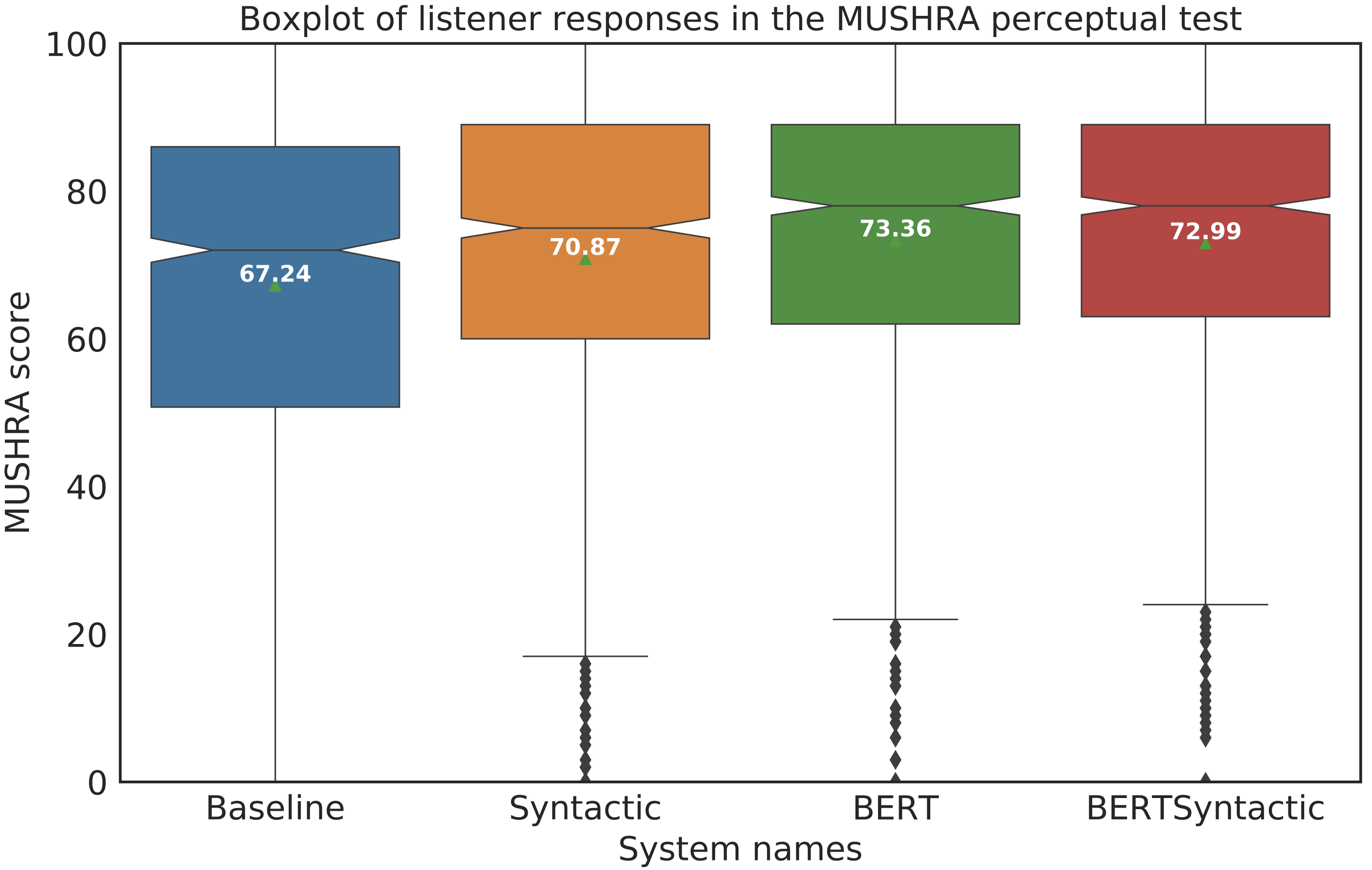}
	\captionsetup{justification=centering}
	\caption{Prosody MUSHRA results on CPE dataset.}
\vspace{-2mm}
	\label{fig:cpe-ads-mushra}
\end{figure}

Figure~\ref{fig:cpe-ads-mushra} and Table~\ref{tab:cpe} report the MUSHRA scores for evaluating prosody and naturalness respectively for the test systems on the CPE dataset. These results answer \emph{\textbf{Q2}}, as the proposed approach improves significantly over the baseline on both grounds. It gives us evidence to support our hypothesis that linguistics-driven acoustic embedding selection can significantly improve speech quality. We also observe that better prosody does not translate into improved naturalness and that there is a need to improve acoustic modeling in order to better reflect the prosodic improvements achieved.

\begin{table}[h!]
\begin{center}
\captionsetup{justification=centering}
\caption{\label{tab:cpe} Naturalness MUSHRA on CPE dataset. Fields in bold are indicative of best results. * depicts statistical insignificance in comparison to baseline (p$>$0.05)}
\scalebox{0.9}{
\begin{tabular}{{|l|c|c|c|c|}}
\hline \bf  & \bf Baseline & \bf Syntactic & \bf BERT & \bf BERT\\ & & & & \bf Syntactic \ \\ \hline
Naturalness & 61.84 & 61.36* & 63.67*  & \bf 64.0  \\
\hline
\end{tabular}}
\end{center}
\vspace{-2mm}
\end{table}

We validate the differences between MUSHRA scores using pairwise t-test. All systems improved significantly over the baseline prosody (p$<$0.01). For naturalness, only BERT Syntactic improved over the baseline significantly (p=0.04).

\begin{table}[h!]
\begin{center}
\captionsetup{justification=centering}
\caption{Prosody evaluation breakdown by categories on CPE}
\label{tab:cpesplit}
\scalebox{0.85}{
\begin{tabular}{{|c|c|c|c|c|}}
\hline  \bf System & \bf `wh'  & \bf `or' & \bf compound & \bf Complex \\
& \bf questions & \bf questions & \bf nouns & \\ \hline
Baseline &  64.35 & 56.89 & 70.05 & 68.84 \\
Syntactic & 68.39 & 66.04 & 70.46  & 71.36 \\
BERT & 71.26 & 73.15 & \bf 71.25  & \bf 75.05 \\
BERT Syntactic & \bf 72.25 & \bf 78.15 & 70.13 & 73.69\\
\hline
\end{tabular}}
\end{center}
\vspace{-2mm}
\end{table}

\emph{\textbf{Q3}} is explored in Table~\ref{tab:cpesplit}, which gives the breakdown of prosody results by major categories in CPE. For `wh' questions, we observe that Syntactic alone brings an improvement of 4\% and BERT Syntactic performs the best by improving 8\% over the baseline. This suggests that `wh' questions  generally share a closely related syntax structure and that information can be used to achieve better prosody. This intuition is further strengthened by the improvements observed for `or' questions. Syntactic alone improves by 9\% over the baseline and BERT Syntactic performs the best by improving 21\% over the baseline. The improvement observed in `or' questions is greater than `wh' questions as most `or' questions have a syntax structure unique to them. For both categories, the systems Syntactic, BERT and BERT Syntactic show incremental improvement. Thus, it is evident that the extent of syntactic information captured drives speech synthesis quality for these two categories.

Compound nouns proved harder to improve upon as compared to questions. BERT performed the best in this category with a 1.2\% absolute improvement over the baseline. We speculate that BERT’s ability to encode semantic information in additional to distributional one is crucial in its better treatment of compounds. The stress pattern of nominal compounds crucially depends on the semantics of the entities involved.

For other complex sentences, BERT performed the best by improving 6\% over the baseline. This can be attributed to the fact that most of the complex sentences required contextual knowledge. Although Syntactic does improve over the baseline, syntax does not look like the driving factor as BERT Syntactic performs a bit worse than BERT. This indicates that enhancing syntax representation hinders BERT from fully leveraging the contextual knowledge it captured to drive embedding selection. 

\begin{figure}
	\centering	\includegraphics[width=0.95\linewidth]{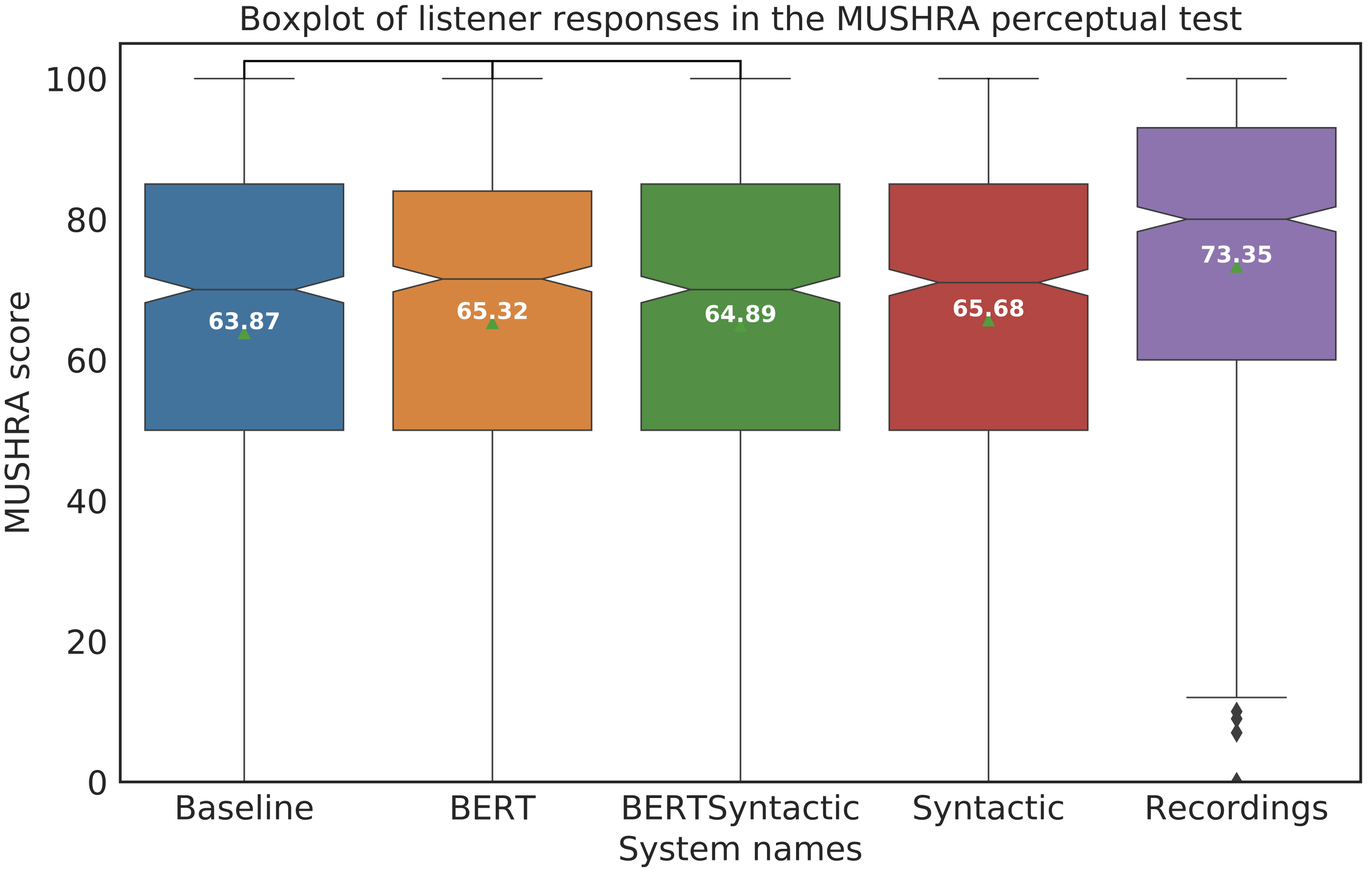}
	\captionsetup{justification=centering}
	\caption{Naturalness MUSHRA results on LFR dataset. Joining line between two systems signifies statistical insignificance in comparison to baseline(p$>$0.05)}
\vspace{-2mm}
	\label{fig:lfr-mushra}
\end{figure}

\emph{\textbf{Q4}} is answered in Figure~\ref{fig:lfr-mushra}, which reports the MUSHRA scores on the LFR dataset. Only the Syntactic system improved over baseline with statistical significance (p=0.02). We close the gap between the baseline and the recordings by almost 20\%.  

To achieve suitable prosody, LFR requires longer distance dependencies and knowledge of prosodic groups. Such information can be approximated more effectively by the Syntactic system rather than the CWE based systems. 

\section{Conclusions}
\vspace{-1mm}
\label{sec:page}
Current VAE-based TTS systems are susceptible to monotonous speech generation due to the need to select a suitable acoustic embedding for synthesis. In this work, we propose a novel approach to leverage linguistic information to drive the embedding selection of such systems. We demonstrate, that doing so is better than simply using the information as modeling features. 

Our approach improves the generated speech in both prosody and naturalness. We propose 3 systems (Syntactic, BERT and BERT Syntactic) and evaluated their performance on 2 datasets: common prosodic errors and LFR. The Syntactic system improved significantly over the baseline on most parameters. Information captured by BERT further improved prosody for cases where contextual knowledge was required. For LFR, we bridged the gap between baseline and recordings by \texttildelow20\%. 

\bibliographystyle{IEEEtran}

\bibliography{template}

\begin{thebibliography}{10}
\providecommand{\url}[1]{#1}
\csname url@samestyle\endcsname
\providecommand{\newblock}{\relax}
\providecommand{\bibinfo}[2]{#2}
\providecommand{\BIBentrySTDinterwordspacing}{\spaceskip=0pt\relax}
\providecommand{\BIBentryALTinterwordstretchfactor}{4}
\providecommand{\BIBentryALTinterwordspacing}{\spaceskip=\fontdimen2\font plus
\BIBentryALTinterwordstretchfactor\fontdimen3\font minus
  \fontdimen4\font\relax}
\providecommand{\BIBforeignlanguage}[2]{{%
\expandafter\ifx\csname l@#1\endcsname\relax
\typeout{** WARNING: IEEEtran.bst: No hyphenation pattern has been}%
\typeout{** loaded for the language `#1'. Using the pattern for}%
\typeout{** the default language instead.}%
\else
\language=\csname l@#1\endcsname
\fi
#2}}
\providecommand{\BIBdecl}{\relax}
\BIBdecl

\bibitem{shen2018natural}
J.~Shen, R.~Pang, R.~J. Weiss, M.~Schuster, N.~Jaitly, Z.~Yang, Z.~Chen,
  Y.~Zhang, Y.~Wang, R.~Skerrv-Ryan \emph{et~al.}, ``Natural tts synthesis by
  conditioning wavenet on mel spectrogram predictions,'' in \emph{2018 IEEE
  International Conference on Acoustics, Speech and Signal Processing
  (ICASSP)}.\hskip 1em plus 0.5em minus 0.4em\relax IEEE, 2018, pp. 4779--4783.

\bibitem{prenger2019waveglow}
R.~Prenger, R.~Valle, and B.~Catanzaro, ``Waveglow: A flow-based generative
  network for speech synthesis,'' in \emph{ICASSP 2019-2019 IEEE International
  Conference on Acoustics, Speech and Signal Processing (ICASSP)}.\hskip 1em
  plus 0.5em minus 0.4em\relax IEEE, 2019, pp. 3617--3621.

\bibitem{lorenzo2019towards}
J.~Lorenzo-Trueba, T.~Drugman, J.~Latorre, T.~Merritt, B.~Putrycz,
  R.~Barra-Chicote, A.~Moinet, and V.~Aggarwal, ``Towards achieving robust
  universal neural vocoding,'' in \emph{Proc. INTERSPEECH 2019}, 09 2019, pp.
  181--185.

\bibitem{hsu2018hierarchical}
\BIBentryALTinterwordspacing
W.-N. Hsu, Y.~Zhang, R.~Weiss, H.~Zen, Y.~Wu, Y.~Cao, and Y.~Wang,
  ``Hierarchical generative modeling for controllable speech synthesis,'' in
  \emph{International Conference on Learning Representations}, 2019. [Online].
  Available: \url{https://openreview.net/forum?id=rygkk305YQ}
\BIBentrySTDinterwordspacing

\bibitem{govender2018using}
A.~Govender and S.~King, ``Using pupillometry to measure the cognitive load of
  synthetic speech,'' \emph{System}, vol.~50, p. 100.

\bibitem{akuzawa2018expressive}
\BIBentryALTinterwordspacing
K.~Akuzawa, Y.~Iwasawa, and Y.~Matsuo, ``Expressive speech synthesis via
  modeling expressions with variational autoencoder,'' in \emph{Proc.
  Interspeech 2018}, 2018, pp. 3067--3071. [Online]. Available:
  \url{http://dx.doi.org/10.21437/Interspeech.2018-1113}
\BIBentrySTDinterwordspacing

\bibitem{SyntaxProsody}
\BIBentryALTinterwordspacing
A.~Köhn, T.~Baumann, and O.~Dörfler, ``An empirical analysis of the
  correlation of syntax and prosody,'' in \emph{Proc. INTERSPEECH 2018}, 2018,
  pp. 2157--2161. [Online]. Available:
  \url{http://dx.doi.org/10.21437/Interspeech.2018-2530}
\BIBentrySTDinterwordspacing

\bibitem{ProsodyReview}
\BIBentryALTinterwordspacing
M.~Wagner and D.~G. Watson, ``Experimental and theoretical advances in prosody:
  A review,'' \emph{Language and Cognitive Processes}, vol.~25, no. 7-9, pp.
  905--945, 2010, pMID: 22096264. [Online]. Available:
  \url{https://doi.org/10.1080/01690961003589492}
\BIBentrySTDinterwordspacing

\bibitem{Guo2019}
\BIBentryALTinterwordspacing
H.~Guo, F.~K. Soong, L.~He, and L.~Xie, ``Exploiting syntactic features in a
  parsed tree to improve end-to-end tts,'' in \emph{Proc. INTERSPEECH 2019},
  2019, pp. 4460--4464. [Online]. Available:
  \url{http://dx.doi.org/10.21437/Interspeech.2019-2167}
\BIBentrySTDinterwordspacing

\bibitem{Aubin2019}
\BIBentryALTinterwordspacing
A.~Aubin, A.~Cervone, O.~Watts, and S.~King, ``Improving speech synthesis with
  discourse relations,'' in \emph{Proc. INTERSPEECH 2019}, 2019, pp.
  4470--4474. [Online]. Available:
  \url{http://dx.doi.org/10.21437/Interspeech.2019-1945}
\BIBentrySTDinterwordspacing

\bibitem{tenney}
\BIBentryALTinterwordspacing
I.~Tenney, P.~Xia, B.~Chen, A.~Wang, A.~Poliak, R.~T. McCoy, N.~Kim, B.~V.
  Durme, S.~R. Bowman, D.~Das, and E.~Pavlick, ``What do you learn from
  context? probing for sentence structure in contextualized word
  representations,'' in \emph{International Conference on Learning
  Representations}, 2019. [Online]. Available:
  \url{https://openreview.net/forum?id=SJzSgnRcKX}
\BIBentrySTDinterwordspacing

\bibitem{mikolov2013distributed}
\BIBentryALTinterwordspacing
T.~Mikolov, I.~Sutskever, K.~Chen, G.~S. Corrado, and J.~Dean, ``Distributed
  representations of words and phrases and their compositionality,'' in
  \emph{Advances in Neural Information Processing Systems 26}, C.~J.~C. Burges,
  L.~Bottou, M.~Welling, Z.~Ghahramani, and K.~Q. Weinberger, Eds., 2013, pp.
  3111--3119. [Online]. Available:
  \url{http://papers.nips.cc/paper/5021-distributed-representations-of-words-and-phrases-and-their-compositionality.pdf}
\BIBentrySTDinterwordspacing

\bibitem{CWEEffects}
\BIBentryALTinterwordspacing
S.~Stehwien, N.~T. Vu, and A.~Schweitzer, ``Effects of word embeddings on
  neural network-based pitch accent detection,'' in \emph{Proc. 9th
  International Conference on Speech Prosody 2018}, 2018, pp. 719--723.
  [Online]. Available: \url{http://dx.doi.org/10.21437/SpeechProsody.2018-146}
\BIBentrySTDinterwordspacing

\bibitem{stanton2018predicting}
D.~Stanton, Y.~Wang, and R.~Skerry-Ryan, ``Predicting expressive speaking style
  from text in end-to-end speech synthesis,'' in \emph{2018 IEEE Spoken
  Language Technology Workshop (SLT)}.\hskip 1em plus 0.5em minus 0.4em\relax
  IEEE, 2018, pp. 595--602.

\bibitem{semeval}
\BIBentryALTinterwordspacing
D.~Cer, M.~Diab, E.~Agirre, I.~Lopez-Gazpio, and L.~Specia, ``{S}em{E}val-2017
  task 1: Semantic textual similarity multilingual and crosslingual focused
  evaluation,'' in \emph{Proceedings of the 11th International Workshop on
  Semantic Evaluation ({S}em{E}val-2017)}.\hskip 1em plus 0.5em minus
  0.4em\relax Vancouver, Canada: Association for Computational Linguistics,
  Aug. 2017, pp. 1--14. [Online]. Available:
  \url{https://www.aclweb.org/anthology/S17-2001}
\BIBentrySTDinterwordspacing

\bibitem{Dall2016}
R.~Dall, K.~Hashimoto, K.~Oura, Y.~Nankaku, and K.~Tokuda, ``Redefining the
  linguistic context feature set for hmm and dnn tts through position and
  parsing,'' in \emph{INTERSPEECH}, 2016.

\bibitem{shen2017neural}
\BIBentryALTinterwordspacing
Y.~Shen, Z.~Lin, C.~wei Huang, and A.~Courville, ``Neural language modeling by
  jointly learning syntax and lexicon,'' in \emph{International Conference on
  Learning Representations}, 2018. [Online]. Available:
  \url{https://openreview.net/forum?id=rkgOLb-0W}
\BIBentrySTDinterwordspacing

\bibitem{SyntacticDistance}
\BIBentryALTinterwordspacing
Y.~Shen, Z.~Lin, A.~P. Jacob, A.~Sordoni, A.~Courville, and Y.~Bengio,
  ``Straight to the tree: Constituency parsing with neural syntactic
  distance,'' in \emph{Proceedings of the 56th Annual Meeting of the
  Association for Computational Linguistics (Volume 1: Long Papers)}.\hskip 1em
  plus 0.5em minus 0.4em\relax Melbourne, Australia: Association for
  Computational Linguistics, Jul. 2018, pp. 1171--1180. [Online]. Available:
  \url{https://www.aclweb.org/anthology/P18-1108}
\BIBentrySTDinterwordspacing

\bibitem{BERT}
\BIBentryALTinterwordspacing
J.~Devlin, M.-W. Chang, K.~Lee, and K.~Toutanova, ``{BERT}: Pre-training of
  deep bidirectional transformers for language understanding,'' in
  \emph{Proceedings of the 2019 Conference of the North {A}merican Chapter of
  the Association for Computational Linguistics: Human Language Technologies,
  Volume 1 (Long and Short Papers)}.\hskip 1em plus 0.5em minus 0.4em\relax
  Minneapolis, Minnesota: Association for Computational Linguistics, Jun. 2019,
  pp. 4171--4186. [Online]. Available:
  \url{https://www.aclweb.org/anthology/N19-1423}
\BIBentrySTDinterwordspacing

\bibitem{wordpiece}
M.~Schuster and K.~Nakajima, ``Japanese and korean voice search,'' in
  \emph{International Conference on Acoustics, Speech and Signal Processing},
  2012, pp. 5149--5152.

\bibitem{Tacotron}
Y.~Wang, R.~J. Skerry-Ryan, D.~Stanton, Y.~Wu, R.~J. Weiss, N.~Jaitly, Z.~Yang,
  Y.~Xiao, Z.~Chen, S.~Bengio, Q.~V. Le, Y.~Agiomyrgiannakis, R.~Clark, and
  R.~A. Saurous, ``Tacotron: Towards end-to-end speech synthesis,'' in
  \emph{INTERSPEECH}, 2017.

\bibitem{prateek2019other}
N.~Prateek, M.~{\L}ajszczak, R.~Barra-Chicote, T.~Drugman, J.~Lorenzo-Trueba,
  T.~Merritt, S.~Ronanki, and T.~Wood, ``In other news: A bi-style
  text-to-speech model for synthesizing newscaster voice with limited data,''
  \emph{NAACL HLT 2019}, pp. 205--213, 2019.

\bibitem{latorre2019effect}
J.~Latorre, J.~Lachowicz, J.~Lorenzo-Trueba, T.~Merritt, T.~Drugman,
  S.~Ronanki, and V.~Klimkov, ``Effect of data reduction on
  sequence-to-sequence neural tts,'' in \emph{ICASSP 2019-2019 IEEE
  International Conference on Acoustics, Speech and Signal Processing
  (ICASSP)}.\hskip 1em plus 0.5em minus 0.4em\relax IEEE, 2019, pp. 7075--7079.

\bibitem{hodari2019using}
\BIBentryALTinterwordspacing
Z.~Hodari, O.~Watts, and S.~King, ``Using generative modelling to produce
  varied intonation for speech synthesis,'' in \emph{Proc. 10th ISCA Speech
  Synthesis Workshop}, 2019, pp. 239--244. [Online]. Available:
  \url{http://dx.doi.org/10.21437/SSW.2019-43}
\BIBentrySTDinterwordspacing

\bibitem{CPEDataset}
\BIBentryALTinterwordspacing
W.~Coster and D.~Kauchak, ``Learning to simplify sentences using {W}ikipedia,''
  in \emph{Proceedings of the Workshop on Monolingual Text-To-Text
  Generation}.\hskip 1em plus 0.5em minus 0.4em\relax Portland, Oregon:
  Association for Computational Linguistics, Jun. 2011, pp. 1--9. [Online].
  Available: \url{https://www.aclweb.org/anthology/W11-1601}
\BIBentrySTDinterwordspacing

\bibitem{GoogleLFR}
\BIBentryALTinterwordspacing
R.~Clark, H.~Silen, T.~Kenter, and R.~Leith, ``Evaluating long-form
  text-to-speech: Comparing the ratings of sentences and paragraphs,'' in
  \emph{Proc. 10th ISCA Speech Synthesis Workshop}, 2019, pp. 99--104.
  [Online]. Available: \url{http://dx.doi.org/10.21437/SSW.2019-18}
\BIBentrySTDinterwordspacing

\bibitem{recommendation2001method}
I.~Recommendation, ``Method for the subjective assessment of intermediate sound
  quality (mushra),'' \emph{ITU, BS}, pp. 1543--1, 2001.

\end{thebibliography}


\end{document}